\documentclass[conference]{IEEEtran}
\IEEEoverridecommandlockouts
\usepackage{cite}
\usepackage{amsmath,amssymb,amsfonts}
\usepackage{algorithmic}
\usepackage{graphicx}
\usepackage{textcomp}
\usepackage{xcolor}
\usepackage{float}
\usepackage{booktabs}
\usepackage{multicol,multirow}
\usepackage[nolist]{acronym}
\usepackage{balance}
\usepackage{array}
\usepackage[caption=false]{subfig}
\def\BibTeX{{\rm B\kern-.05em{\sc i\kern-.025em b}\kern-.08em
    T\kern-.1667em\lower.7ex\hbox{E}\kern-.125emX}}

\usepackage{todonotes}

\newcolumntype{L}[1]{>{\raggedright\let\newline\\\arraybackslash}m{#1}}
\newcolumntype{C}[1]{>{\centering\let\newline\\\arraybackslash}m{#1}}
\newcolumntype{R}[1]{>{\raggedleft\let\newline\\\arraybackslash}m{#1}}

\begin{acronym}[TDMA]
    \acro{AUC-PR}{Area Under the Precision-Recall Curve}
    \acro{AUC-ROC}{Area Under the ROC Curve}
    \acro{CNN}{Convolutional Neural Network}
    \acro{HOG}{Histogram of Oriented Gradients}
    \acro{LR}{Logistic Regression}
    \acro{LBP}{Local Binary Patterns}
    \acro{LPQ}{Local Phase Quantization}
    \acro{SURF}{Speeded Up Robust Features}
    \acro{SVM}{Support Vector Machine}
\end{acronym}

\newif\ifauthors
\authorstrue
    
\begin{document}

\title{Deep Single Models vs. Ensembles: Insights for a Fast Deployment of Parking Monitoring Systems}

\ifauthors
\author{
    \IEEEauthorblockN{Andre Gustavo Hochuli$^*$\\Jean Paul Barddal}
	\IEEEauthorblockA{Graduate Program in Informatics (PPGIa)\\
	    Pontif\'icia Universidade Cat\'olica do Paran\'a\\
		Curitiba, PR - Brazil\\
		\{aghochuli, jpbarddal\}@ppgia.pucpr.br
        }
            
    \and

    \IEEEauthorblockN{Gillian Cezar Palhano\\Leonardo Matheus Mendes}
	\IEEEauthorblockA{
	    Pontif\'icia Universidade Cat\'olica do Paran\'a\\
		Curitiba, PR - Brazil\\		
            gillian.palhano@pucpr.edu.br\\
            l.mendes2@pucpr.edu.br}   
    
    \and    
    
    \IEEEauthorblockN{Paulo  Ricardo Lisboa de Almeida}
	\IEEEauthorblockA{Department of Informatics (DInf)\\
	    Universidade Federal do Paran\'a\\
		Curitiba, PR - Brazil\\
		paulorla@ufpr.br}}
\else 
\author{
    \IEEEauthorblockN{Anonymous Authors}
    \vspace*{2.5cm}
}
\fi
\maketitle

\begin{abstract}
Searching for available parking spots in high-density urban centers is a stressful task for drivers that can be mitigated by systems that know in advance the nearest parking space available.
To this end, image-based systems offer cost advantages over other sensor-based alternatives (e.g., ultrasonic sensors), requiring less physical infrastructure for installation and maintenance.
Despite recent deep learning advances, deploying intelligent parking monitoring is still a challenge since most approaches involve collecting and labeling large amounts of data, which is laborious and time-consuming. Our study aims to uncover the challenges in creating a global framework, trained using publicly available labeled parking lot images, that performs accurately across diverse scenarios, enabling the parking space monitoring as a ready-to-use system to deploy in a new environment. Through exhaustive experiments involving different datasets and deep learning architectures, including fusion strategies and ensemble methods, we found that models trained on diverse datasets can achieve 95\% accuracy without the burden of data annotation and model training on the target parking lot.
\end{abstract}

\begin{IEEEkeywords}
Parking Lot Monitoring, Parking Space Classification, Holistic Classification Models
\end{IEEEkeywords}

\section{Introduction}\label{sec:intro}

Searching for vacant parking spots in high-density urban centers is a common issue. Consequently, an efficient parking lot system is needed to assist drivers in swiftly and conveniently parking their cars. In this context, image-based approaches are a common choice to determine parking spot occupancy due to their cost advantages over other sensor-based alternatives, often requiring less physical infrastructure for installation and maintenance \cite{paidiEtAl2018}.

Additionally, camera-based systems are well-suited for short-term needs, such as public events, where parking monitoring is necessary for only a few days. In such a case, an initial demarcation of parking spots is required only once. Subsequently, each parking space is cropped from the entire image, and then a model classifies it as occupied or empty.

Recent deep learning advances \cite{grbicKoch2023,vargheseSreelekha2019,dhuriEtAL2021} have shown image-based parking spot classification rates of over 95\% \cite{almeidaEtAl2022} using models tailored for specific parking lot scenarios. Consequently, several approaches and datasets have been released, featuring distinct challenges, diverse amounts of annotated data, and variations in camera angles, weather conditions, and backgrounds.
Nevertheless, most approaches still rely on time-consuming tasks, including data collection, annotation, and model construction. Furthermore, whether an environmental change occurs, such as changes in camera positioning or occlusions, reworking for all these laborious tasks is often necessary. The authors in \cite{almeidaEtAl2022,HochuliEtAl2022-Annotation} concluded that training or fine-tuning models on a target dataset remains a bottleneck yet to be addressed.

Our work targets an analysis of an off-the-shelf solution. The challenge is to accurately predict whether parking spots are free or occupied in a given target parking lot without needing labeled training samples from the target parking lot (i.e., a global model). To this end, we define two research questions to guide such analysis:

\begin{itemize}
    \item RQ1: How accurate are existing deep learning models when applied in a cross-dataset scenario?
    \item RQ2: Regarding different architectures and ensemble strategies, which framework is the most suited for cross-dataset scenarios?
\end{itemize}



    

To address these questions, we conducted exhaustive experiments, considering various state-of-the-art datasets and deep learning architectures. Besides, fusion strategies and ensemble methods were also assessed. Another contribution of this research is the critical analysis across diverse scenarios, which elucidates the challenges and limitations of constructing a global framework that does not require training samples from the target parking lot.


This work is organized as follows: Section \ref{sec:state-of-art} presents the related works on parking slot classification. 
Section \ref{sec:prob_stat} outlines the problem statement and proposed protocol. 
Section \ref{sec:experiments} describes the conducted experiments and discussions. 
Section \ref{sec:conclusion} presents our findings and provides insights for future research.

\section{Related Works}\label{sec:state-of-art}

This section focuses on related works that aim to create cross-dataset models for parking lot monitoring systems.
In other words, we focus on models that do not require instances from the target parking lot for training (for a broader discussion about state-of-the-art regarding other scenarios, refer to \cite{almeidaEtAl2022}). 
As discussed in \cite{almeidaEtAl2022}, cross-datasets models can ease the deployment of parking lot monitoring systems, as no human labor is required to label instances in the target parking lot.

A seminal work that deals with such a problem using a large-scale dataset is \cite{almeidaEtAl2015}, where the PKLot dataset was proposed. 
In the work, the authors use an ensemble of \ac{SVM} classifiers trained using \ac{LPQ} and \ac{LBP} features.

Similarly, the authors in \cite{vitekMelnicuk2018,moraEtAl2018,vargheseSreelekha2019} used \ac{SVM} classifiers. 
In \cite{vitekMelnicuk2018}, a classifier was trained using a combination of parking angle information and \ac{HOG} features. 
The authors argued that by using \ac{HOG} features, it was possible to deploy the trained model in smart cameras, where the processing power is restricted.
In \cite{moraEtAl2018}, the well-known SIFT features were used to train the \ac{SVM}, while \cite{vargheseSreelekha2019} used a combination of the \ac{SURF} \cite{bayEtAl2008} features and color information to train their model. 
The authors in \cite{moraEtAl2018} also tested an approach trained using a VGG16 network.

In \cite{amatoEtAl2017}, the CNRPark-EXT dataset was proposed, which, alongside the PKLot, became an important dataset for the parking spaces classification problem. In their work, the authors proposed a lightweight network for classifying the parking spaces called mAlexNet.

In the same vein, the authors in \cite{nurullayevLee2019,dhuriEtAL2021,HochuliEtAl2022-Annotation,grbicKoch2023} also proposed deep learning-based approaches to deal with the parking spaces classification problem. The CarNet network was proposed in \cite{nurullayevLee2019}, which is a \ac{CNN}-based method that skips pixels in the convolution kernel. The authors in \cite{dhuriEtAL2021} employed the VGG16 network to classify the parking spaces between occupied and empty. A custom 3-layer \ac{CNN} was used in \cite{HochuliEtAl2022-Annotation}, where the authors tested models trained considering samples extracted using rotated rectangles, bounding boxes, and fixed-size squares. More recently, the authors in \cite{grbicKoch2023} used a ResNet34 network to classify the individual parking spaces.

Table \ref{tab:state_of_the_art} summarizes the results achieved by each author. As observed, the results vary broadly between authors and when considering the best and worst results achieved by each author. The results are not directly comparable since authors may have employed different datasets and experimental protocols.

\begin{table}[htpb]
\caption{Results achieved in related works considering that no training samples from the target parking lot are given.}
\label{tab:state_of_the_art}
\footnotesize
\begin{tabular}{L{65px}lrL{49px}}
\hline
\hline
Authors & Classifier & Accuracy (\%) & Datasets Used\\
\hline
\hline
Almeida et al. 2015 & \ac{SVM} &  84 -- 90\% &  PKLot\\

Vítek and Melničuk 2018 & \ac{SVM} &  83 -- 96\% & PKLot + Private Dataset\\

Mora et al. 2018 & \ac{SVM} & 72 -- 85\% & PKLot\\

Mora et al. 2018 & VGG16 & 90 -- 97\% & PKLot\\

Varghese and Sreelekha 2020 & SVM & 82\% & PKLot + CNRPark\\

Amato et al. 2017 & mAlexNet & 93 -- 98\% & PKLot + CNRPark\\

Nurullayev and Lee 2019 & CarNet & 94 -- 98\%& PKLot + CNRPark\\

Dhuri et al. 2021 & VGG16 & 87\% & PKLot + CNRPark + Private Dataset\\

Hochuli et al. 2022  & Custom CNN & 89 -- 97\% & PKLot\\

Grbić and Koch 2023 & ResNet34 & 92 -- 99\%& PKLot + CNRPark\\
\hline
\textbf{Average} & & \textbf{87 -- 93\%}\\
\hline\hline
\end{tabular}
\vspace{-3mm}
\end{table}

\section{Problem Statement}\label{sec:prob_stat}

Concerning the approaches discussed in Section \ref{sec:state-of-art}, certain aspects need further attention. Assuming that a different camera of the same parking lot mimics a cross-scenario could pose a bias. 
For instance, despite the CNR-EXT Dataset\cite{amatoEtAl2017} having nine cameras, all samples were collected simultaneously, encompassing the same environment, lighting conditions, and noise. 
An analogous issue is observed in the PKLot Dataset\cite{almeidaEtAl2015} when comparing UFPR04 and UFPR05 cameras.

In this study, we investigate strategies to construct a scalable and ready-to-use global framework capable of accurately classifying diverse scenarios without requiring extensive adjustments for deployment in new environments. It is important to note that in such cases, the initial demarcation of parking spot positions is necessary only once, and it can be performed manually or via a vehicle occurrence-based algorithm \cite{grbicKoch2023,almeidaEtAl2023}.

Due to the abundance of parking lot datasets in the state of the art,  a strategy is to combine them to create a cross-dataset environment and evaluate it using an unseen dataset, mimicking real-world deployment.

\begin{figure*}[!ht]
  \centering   
  \subfloat[]{\includegraphics[height=0.24\textheight]{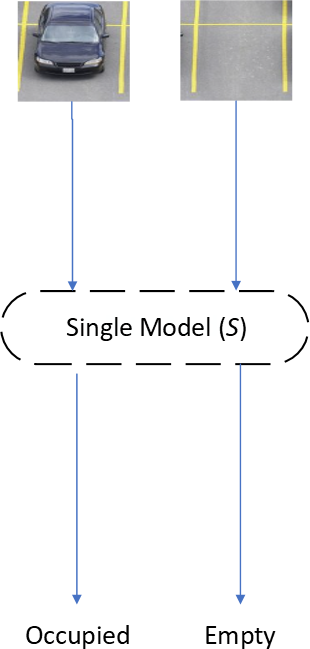}}
  \hfill
  \subfloat[]{\includegraphics[height=0.24\textheight]{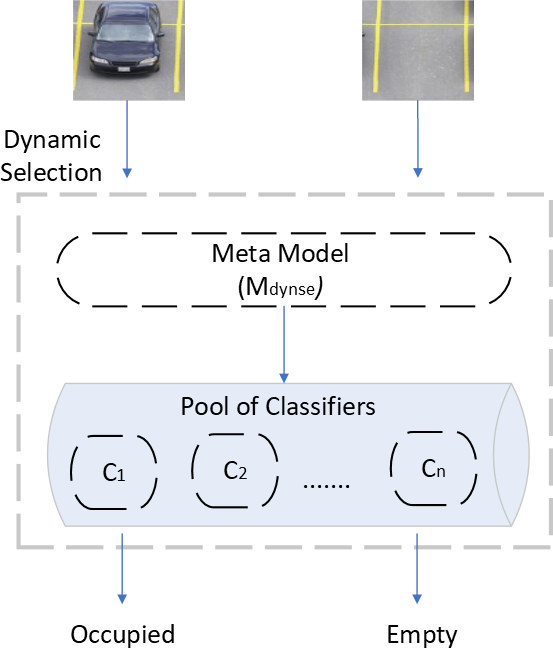}}
  \hfill
  \subfloat[]{\includegraphics[height=0.24\textheight]{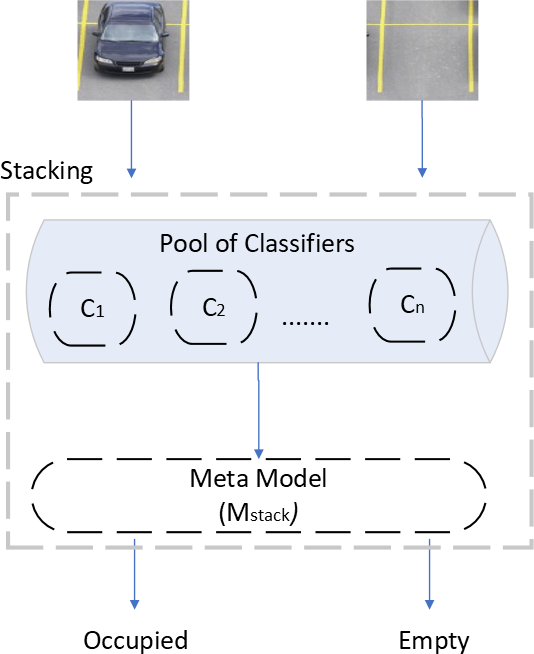}}
  \hfill
  \subfloat[]{\includegraphics[height=0.24\textheight]{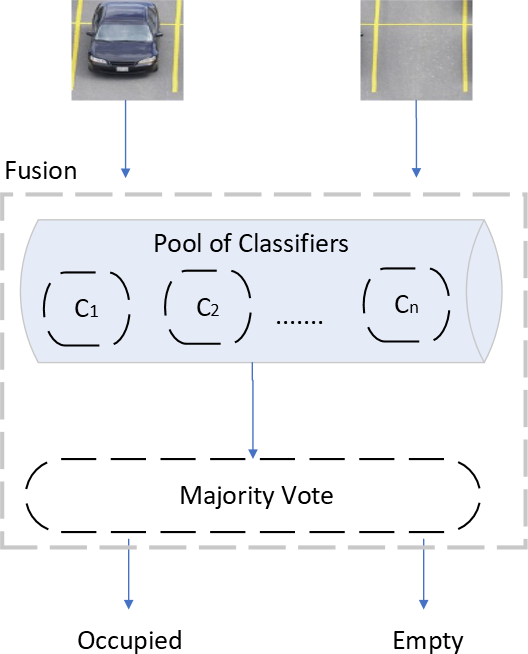}}
  \caption{The proposed strategies assessed for parking lot monitoring systems in cross-dataset scenarios: (a) a single model, and the ensemble-based frameworks named (b) dynamic selection and (c) stacking, and finally (d) the majority vote fusion-based framework.}
  \label{fig:overview_strategies}

\vspace{-1.5mm}
\end{figure*}

To properly answer the proposed research questions (RQ1 and RQ2), four different frameworks are proposed and depicted in Figure \ref{fig:overview_strategies}. First, we deploy an approach concerning a single model ($S$) (Figure \ref{fig:overview_strategies}a). Then, ensemble strategies based on a pool of classifier is presented: Dynamic Selection (Figure \ref{fig:overview_strategies}b) and Stacking (Figure \ref{fig:overview_strategies}c). Finally, the fourth approach (Figure \ref{fig:overview_strategies}d) uses a fusion method based on a majority vote of all individual models. The reasonable here is whether an pool-based method may provide a better generalization against a single global model.

To build a comprehensive analysis concerning model generalization, we also used different architectures to compose the frameworks, as each strategy may provide different representations for the problem, as discussed in Section \ref{sec:experiments}. Further details about the proposed frameworks, classifiers, training protocols, and datasets can be found in Sections \ref{sec:frameworks}, \ref{sec:classifiers}, \ref{sec:training_proto}, and \ref{sec:datasets}, respectively.

\subsection{Proposed Frameworks}\label{sec:frameworks}

This Section describes four proposed approaches to create a global framework for classifying parking lot images in cross-dataset scenarios.

First, a straightforward approach, depicted in Figure \ref{fig:overview_strategies}a, implements a single model ($S$) trained with a combination of diverse scenarios. The reasonable here is to create a robust classifier that will be compared against ensemble strategies. So, the inference is direct: when a parking spot is presented, the model will determine its class. 

A pool of homogeneous classifiers is introduced for the following ensemble-based strategies. In such a case, each individual classifier within the pool ($C_1$, $C_2$, \ldots, $C_n$) is trained in a specific scenario of a parking lot dataset, thereby providing diversity in the pool. Then, a strategy to select the most competent classifier or a fusion scheme is necessary to infer a given input.

Figure \ref{fig:overview_strategies}b illustrates a Dynamic Selection framework where a meta-model ($M_{dynse}$) on the top of the pool learns the competence of each classifier by encoding the feature space of dataset samples. During the inference phase, the model $M_{dynse}$ determines the competent classifier ($C$) based on the input feature space of the test instance.

The Stacking strategy (Figure \ref{fig:overview_strategies}c) relies upon the divergence of classifiers by encoding \textit{a posteriori} probabilities of all models within the pool. A given sample is forwarded through the pool, resulting in a collection of \textit{a posteriori} probabilities provided for each classifier ($C$). Then, a feature vector concatenates each generated \textit{a posteriori} probability. The feature vector encodes the divergence between individual models ($C$) for a given input. So, the meta-model ($M_{stack}$) learns that divergence to provide a class to the input sample.

Finally, as stated in \cite{almeidaEtAl2015}, in the majority vote strategy (Figure \ref{fig:overview_strategies}d), every pool member casts a prediction. That class that receives the absolute majority of votes defines the input class. The advantage of this approach over stacking is its independence from training. However, it does not encode a deep understanding of probability divergences.


\subsection{Classifiers}\label{sec:classifiers}

\begin{figure}[!htb]
  \centering
  \includegraphics[width=0.48\textwidth]{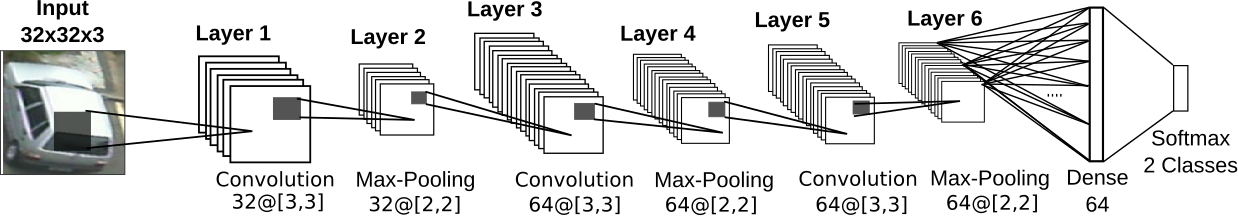}
  \caption{The 3-convolutional layers architecture used to classify parking spots.}
  \label{fig:cnn_arch_pklot}
\vspace{-1mm}
\end{figure}

    
We assessed three different network architectures as the base learner for the Single Model ($S$), the classifiers ($C$) within the classifier pool, and the Meta-Model ($M_{dynse}$), which composes the proposed frameworks depicted in Figure \ref{fig:overview_strategies}.
The first architecture is a convolutional network proposed in \cite{HochuliEtAl2022-Annotation}, tailored to classify parking spots when trained with samples from the target dataset. Figure \ref{fig:cnn_arch_pklot} illustrates this architecture. The model comprises 3-convolutional layers alternating with 2-pooling layers to perform feature extraction. At the end, a dense layer concatenates all features. This short architecture is a cost-effective solution well explored in \cite{HochuliEtAl2022-Annotation}. 

Towards deeper representations, to perform feature extraction, we have used the well-known architectures named MobileNetV3\cite{MobileNetV3} and ResNet-50\cite{ResNet2015}, both pre-trained on Imagenet\cite{ImageNet2015}. To perform classification, we have added two learnable dense layers of sizes 1024 and 128, with ReLU activation to concatenate all features from the frozen convolutional layers. Finally, a softmax activation function provides the class probabilities. The input size is [128,128].

The MobileNet balances accuracy and training complexity. On the other hand, the ResNet-50\cite{ResNet2015}  outperforms various dense network applications with the advantage of residual connections to minimize vanishing gradients at the cost of a high computational power need. Other deep networks, such as VGG16\cite{VGG162014} and EfficientNetB7\cite{EfficientNet2019}, were also evaluated and did not demonstrate improvements over ResNet-50 architecture, and thus, we do not show their results. Table \ref{tab:arch_params} briefly outlines the complexity of each architecture.

\begin{table}[!htb]
\centering
\caption{Architectures Overview}
\label{tab:arch_params}
\begin{tabular}{@{}cccc@{}}
\hline\hline
\multirow{2}{*}{\textbf{Model}} & \multirow{2}{*}{\textbf{Input Size}} & \multicolumn{2}{c}{\textbf{Parameters}} \\ \cmidrule(l){3-4} 
 &  & \textbf{Total} & \textbf{Trainable} \\ 
 \hline \hline
3-Conv.Layers\cite{HochuliEtAl2022-Annotation} & {[}32,32{]} & $\sim$158 K & $\sim$158 K \\ 

MobileNetV3-Large\cite{MobileNetV3} & {[}128,128{]} & $\sim$4.1 M & $\sim$1.2 M \\ 

ResNet-50\cite{ResNet2015} & {[}128,128{]} & $\sim$25 M & $\sim$3.2 M \\ 
\hline\hline
\end{tabular}%
\vspace{-1mm}
\end{table}

Finally, to encode the pool probabilities the meta-model $M_{stack}$ (Figure \ref{fig:overview_strategies}d) are defined as follows: a) a SVM with RBF Kernel and the relaxing parameter $C=0.1$, and b) a shallow MLP composed of three layers of sizes [16, 8, 2].

\subsection{Training Protocol}\label{sec:training_proto} 

All trainable layers are updated with the Adam Optimizer using back-propagation with batches of 64 instances. 
The learning rate is set to $10^{-3}$ initially to allow the weights to quickly fit the long ravines in the weight space, after which it is reduced over time to make the weights fit the sharp curvatures. The network makes use of the well-known cross-entropy loss function. The regularization was implemented through early stopping. The models were trained in an environment with two NVIDIA RTX A5000 GPUs to expedite the training.

\begin{figure}[!htbp]
  \centering    
  \subfloat[]{\includegraphics[height=1cm,width=1.5cm]{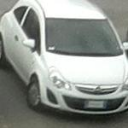}}\hfill      
    \subfloat[]{\includegraphics[height=1cm,width=1.5cm]{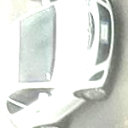}}\hfill
    \subfloat[]{\includegraphics[height=1cm,width=1.5cm]{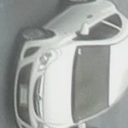}}\hfill
    \subfloat[]{\includegraphics[height=1cm,width=1.5cm]{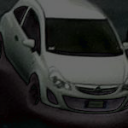}}\hfill
    \subfloat[]{\includegraphics[height=1cm,width=1.5cm]{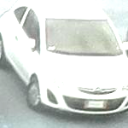}}\hfill
  \caption{The data augmentation enhanced the original representation (a) by mimicking different camera angles (b,c) and lighting conditions (d,e). }
  \label{fig:dataaug}
\vspace{-2mm}
\end{figure}

A synthetic data augmentation was applied for each training batch, improving the generalization through image rotations and changes in contrast and brightness. This approach mimics environmental issues such as different camera angles and lighting conditions. Figure \ref{fig:dataaug} illustrates the resulting images.
\vspace{-2mm}

\subsection{Datasets}\label{sec:datasets}
    
In this work, we used four datasets available in the state-of-the-art, namely PKLot\cite{almeidaEtAl2015}, CNRPark\cite{amatoEtAl2017}, NDISPark\cite{NDIS2022} and BarryStreet\cite{BarryStreetDataset2018}. These datasets provide a wide range of challenges, encompassing variations in camera angles, shadowed cars, weather conditions, environmental settings, and backgrounds, enabling a comprehensive and critical assessment of the proposed frameworks. An important aspect is that they all provide annotations of parking spot locations and binary labels indicating whether it is occupied or empty.

\begin{figure}[htbp]

  \centering    
  \subfloat[]{\includegraphics[height=1.8cm,width=4.0cm,trim={100px 0px 0px 0px}, clip]{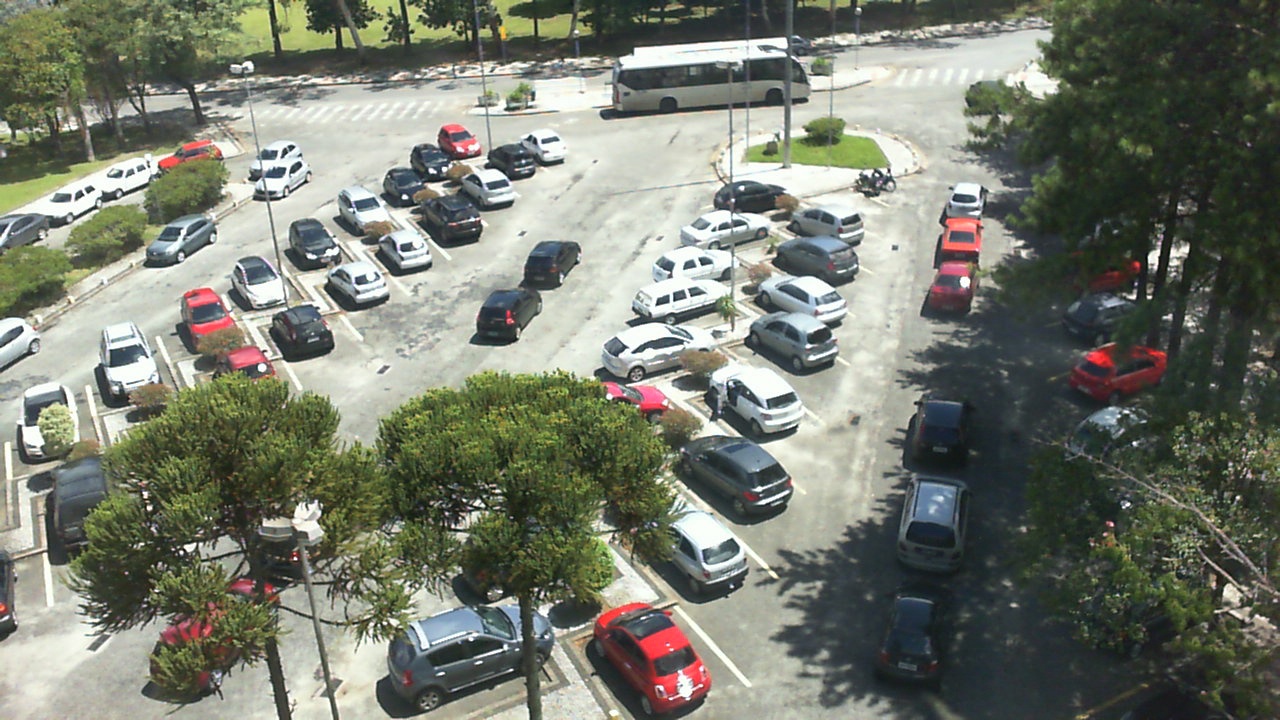}}\hfill
  \subfloat[]{\includegraphics[height=1.8cm,width=4.0cm, trim={100px 0px 0px 0px}, clip]{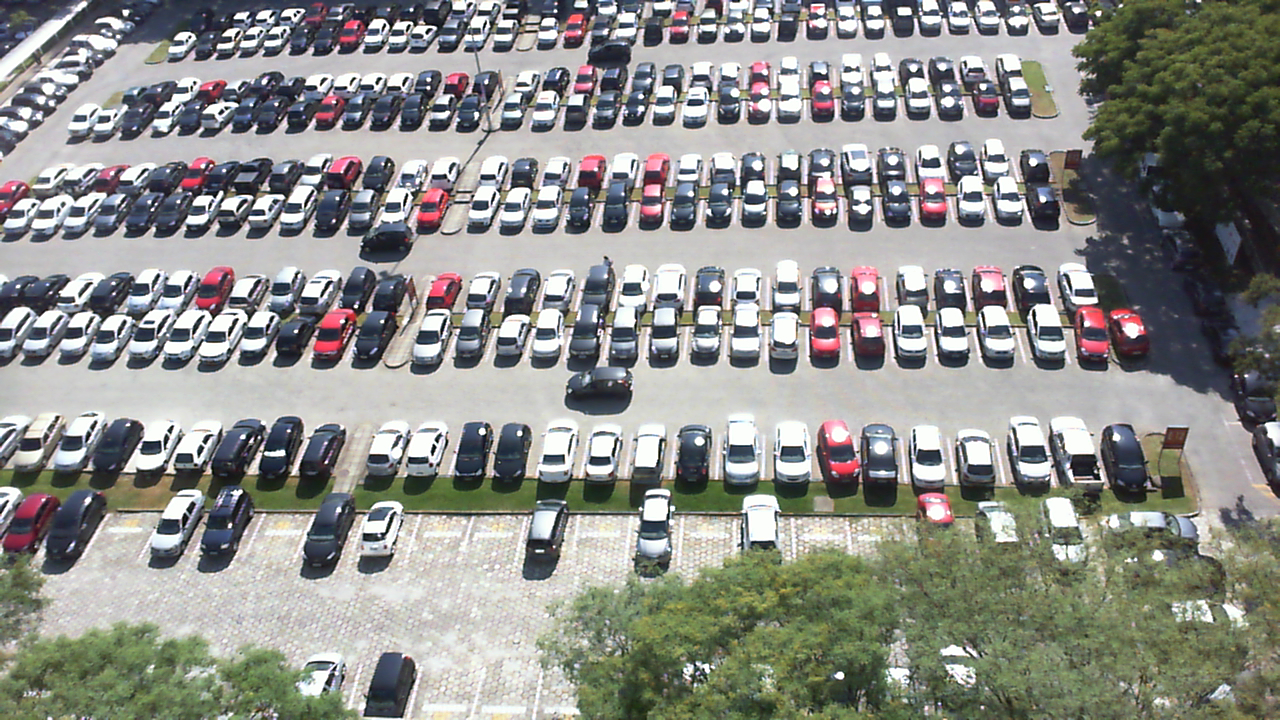}}\quad
  \caption{PKLot dataset image examples.}
  \label{fig:overview_pklot}
\end{figure}

The {PKLot}{\cite{almeidaEtAl2015}} dataset is one of the largest and most comprehensive datasets available for parking lot classification in the state-of-the-art. It includes images captured over about three months, with a time interval of 5 minutes between each image, providing a total of 12,417 parking spots and about 700,000 annotated samples divided into three scenarios named UFPR04, UFPR05, and PUCPR. Image examples of the PKLot dataset are given in Figure \ref{fig:overview_pklot}.

\begin{figure}[htbp]
  \centering    
  \subfloat[]{\includegraphics[height=1.8cm,width=4.0cm,trim={0px 0px 0px 70px}, clip]{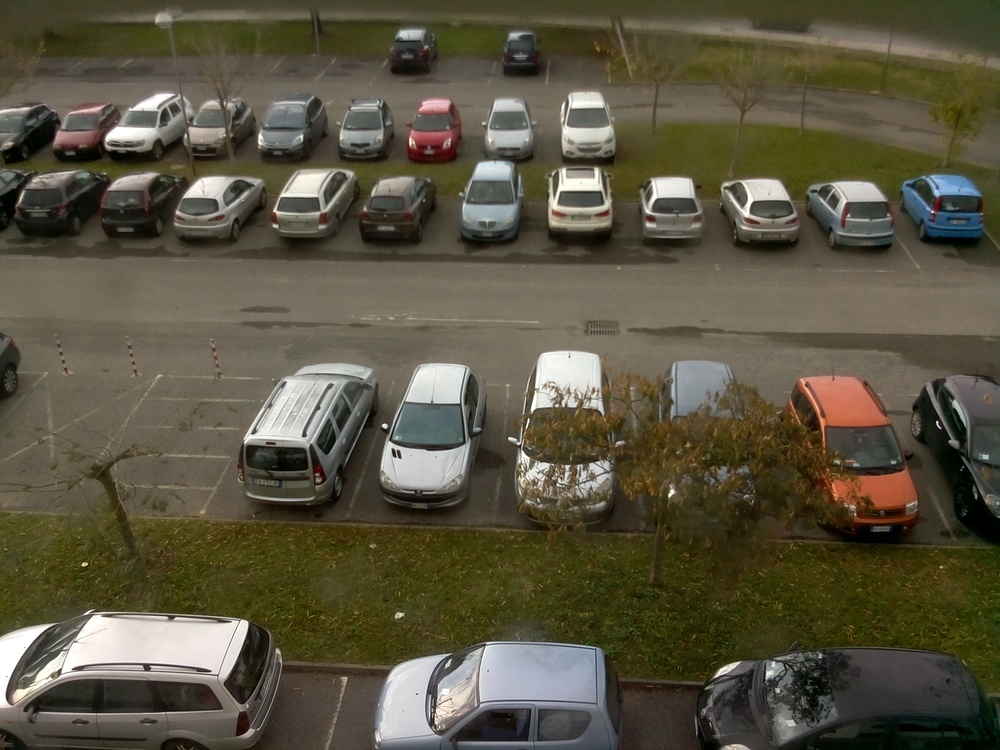}}\hfill
  \subfloat[]{\includegraphics[height=1.8cm,width=4.0cm, trim={0px 0px 0px 70px}, clip]{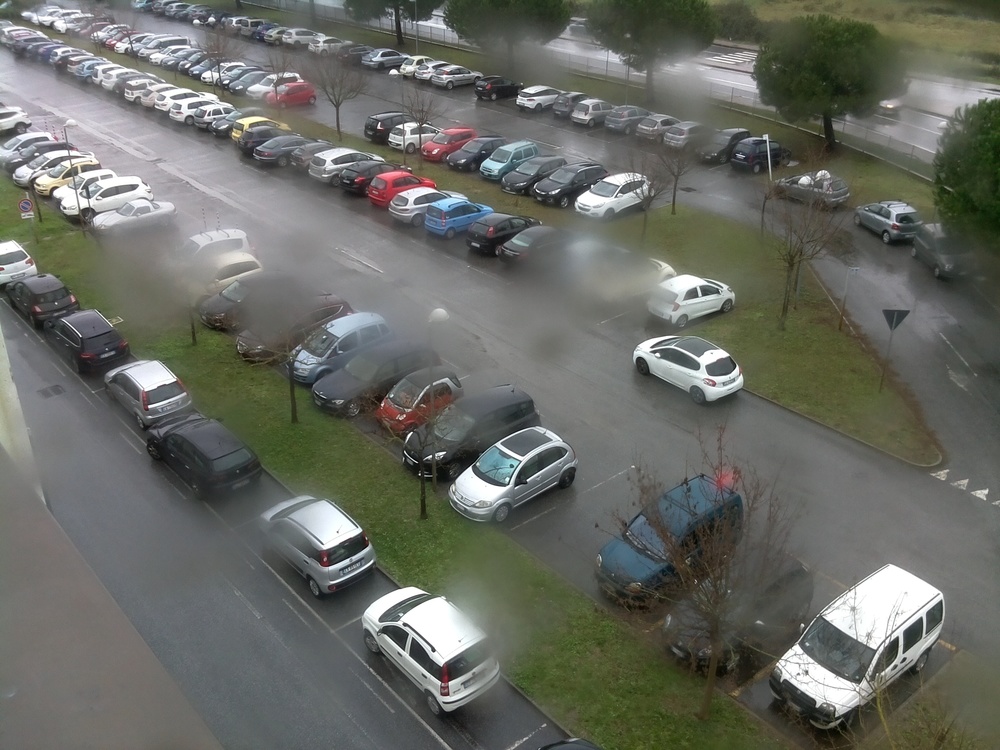}}
    
  \caption{The CNRPark (CNR-EXT) dataset embraces different camera positions for the same environment}
  \label{fig:overview_cnr}
\end{figure}

The CNRPark-EXT \cite{amatoEtAl2017} is another vast dataset containing about 160,000 annotated parking spaces collected from nine cameras. This dataset poses specific challenges, including solar light reflections and raindrops on the camera lens. An example of these challenges is depicted in Figure \ref{fig:overview_cnr}.

\begin{figure}[htbp]
  \centering    
    \subfloat[]{\includegraphics[height=1.8cm,width=4.0cm]{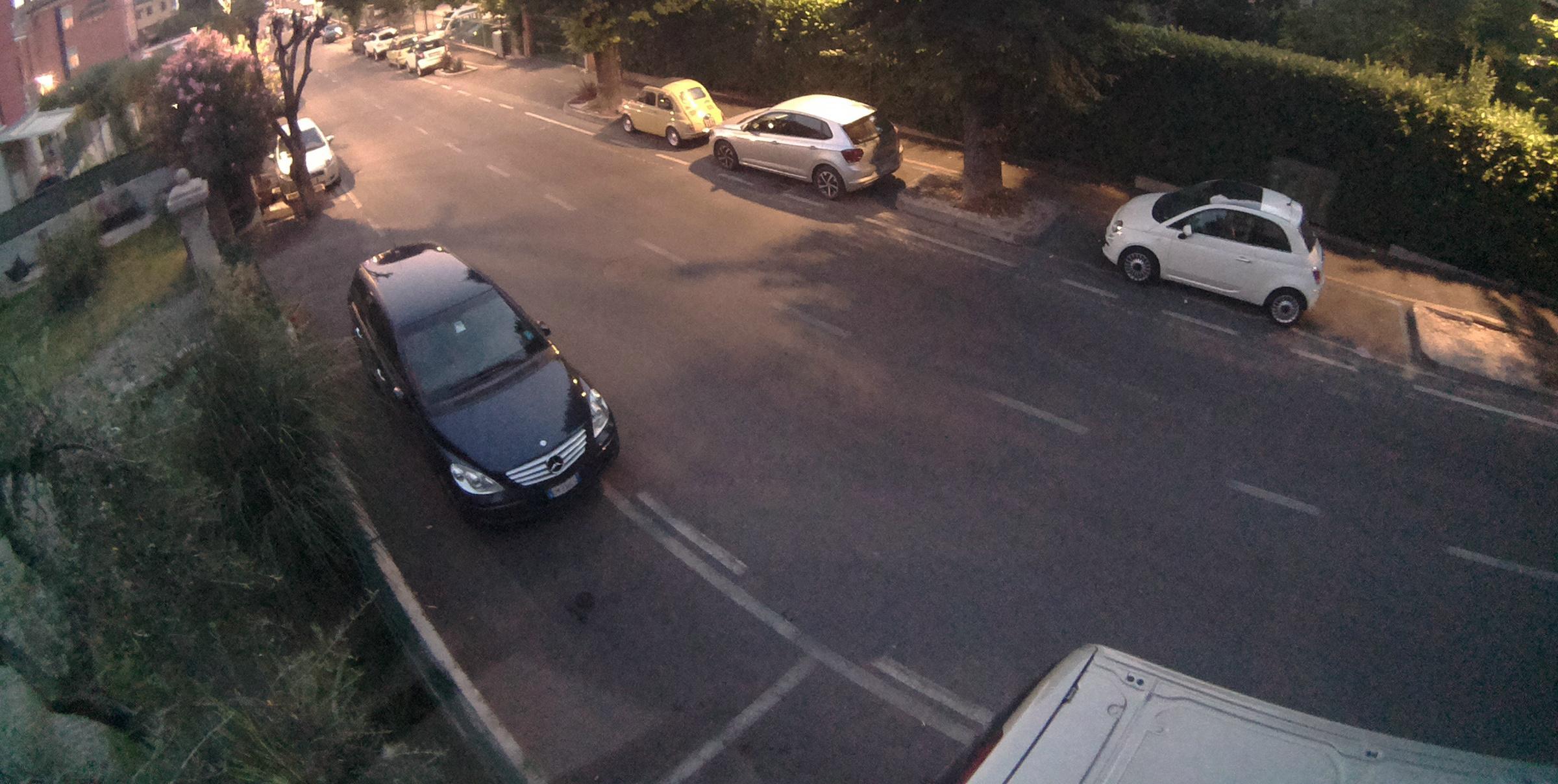}}\hfill
    \subfloat[]{\includegraphics[height=1.8cm,width=4.0cm]{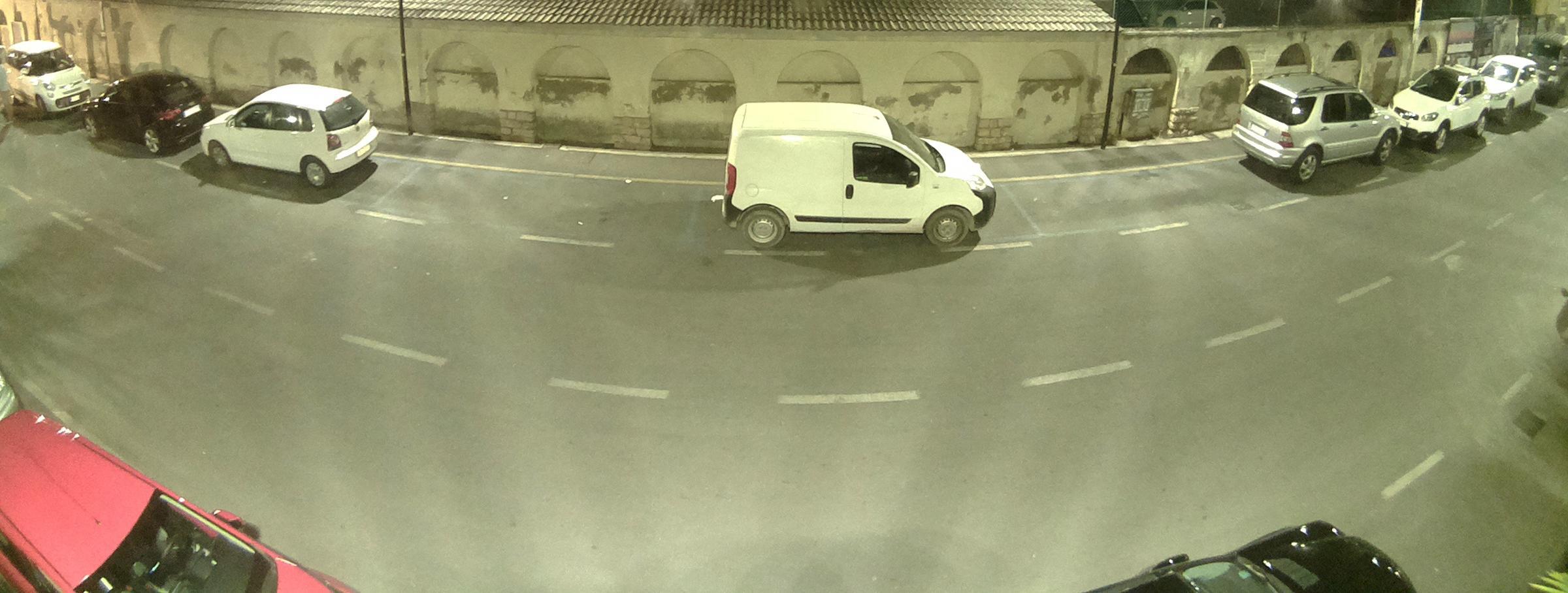}}
  \caption{The NDIS dataset comprises distinct scenarios and challenges, including a lateral view of a street parking lot. }
  \label{fig:overview_ndis}
\end{figure}

The recently released Night and Day Instance Segmented Park Dataset (NDISPark)\cite{NDIS2022} comprises 259 images exhibiting diverse parking areas from seven cameras, encompassing various challenging situations encountered in real scenarios. Additionally, as can be observed in Figure \ref{fig:overview_ndis}, a camera providing a lateral view of street parking spots and partial occlusions caused by obstacles like trees and lampposts contribute to the complexity of the dataset.
\vspace{-5mm}

\begin{figure}[htpb]
  \centering    
  \subfloat[]{\includegraphics[height=1.8cm,width=4.0cm, trim={0px 0px 0px 50px}, clip]{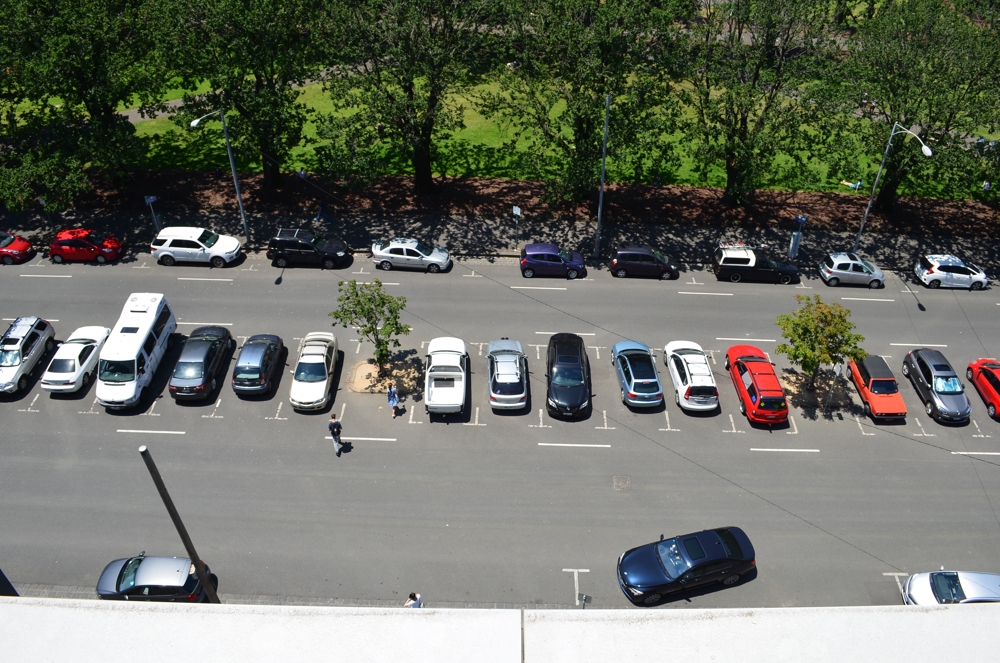}}\hfill
    \subfloat[]{\includegraphics[height=1.8cm,width=4.0cm, trim={0px 0px 0px 50px}, clip]{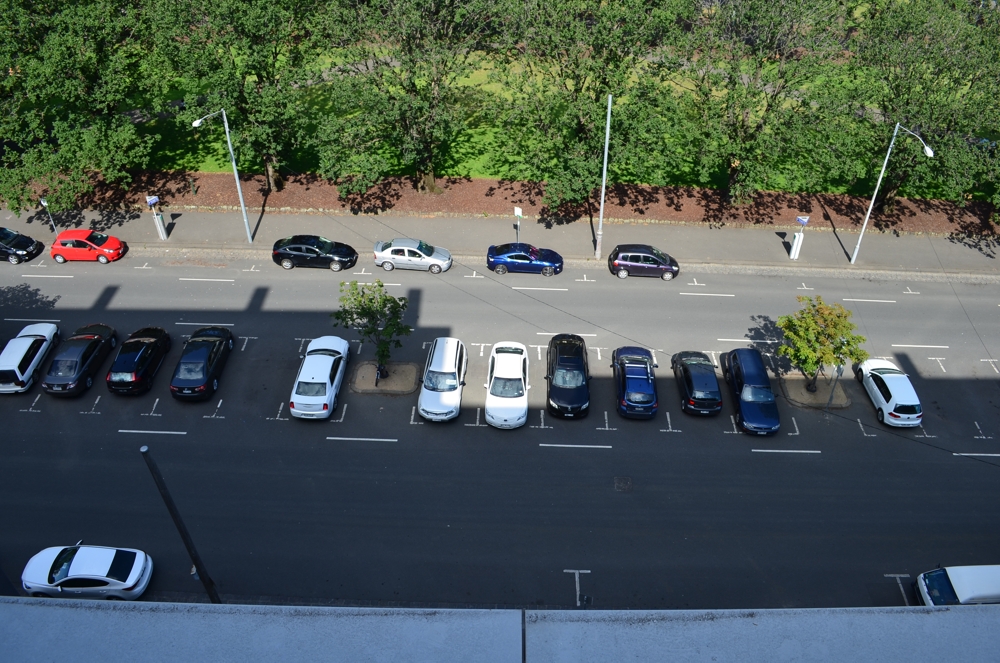}}
    
  \caption{The BarryStreet dataset contains a one-day timelapse of a private parking lot}
  \label{fig:overview_barry}

\end{figure}

The fourth dataset, named Barry Street \cite{BarryStreetDataset2018}, contains images captured at 30-second intervals during daylight hours, offering weather conditions ranging from sunny to cloudy and including shadowed areas. The camera is positioned at a top-view angle, enabling the monitoring of 30 parking spots, as can be seen in Figure \ref{fig:overview_barry}.
\vspace{-5mm}

A description of the amount of data provided from each dataset is presented in Table \ref{tab:parking_spots_per_dataset}. For those interested in a comprehensive description of the datasets, the resulting publications provide in-depth discussions and analysis\cite{almeidaEtAl2022, almeidaEtAl2015, amatoEtAl2017, NDIS2022, BarryStreetDataset2018}. It is worth mentioning that there are other important datasets, such as PLds [9]. However, they contain only bounding boxes surrounding cars, not parking spots.

\begin{table}[!htbp]
\centering
\caption{Parking lots datasets properties}
\label{tab:parking_spots_per_dataset}
\resizebox{1\columnwidth}{!}{
\begin{tabular}{cccccc}
\hline\hline
\textbf{Dataset}  & \textbf{Spots} & \textbf{Days} & \textbf{Cameras} & \textbf{Scenarios} & \textbf{Conditions} \\ 
\hline\hline
PKLot & 695899 & $\sim$90 & 3 & 2 & Day Only\\

CNRPark & 146223 & 23 & 8 & 1 & Day \& Night \\

NDISPark & 2577 & N/D & 7 & 4 & Day \& Night\\

BarryStreet & 2800 & 1 & 1 & 1 & Day Only   \\ 
\hline\hline
                   
\end{tabular}
}
\vspace{-3mm}
\end{table}


\section{Experiments}\label{sec:experiments}

When assessing cross-dataset experiments, one can argue that a common protocol in the state-of-art \cite{almeidaEtAl2022} is training a model in a source scenario and testing in a different target scenario. However, to properly answer our RQ1 and RQ2, we are interested in determining whether a pool of classifiers trained with different scenarios can provide a better generalization.

In this context, we establish two distinct environments. In the first one, we employ the PKLot Dataset\cite{almeidaEtAl2015} for training the frameworks. Therefore, both the Single Model ($S$) and the Meta Classifier ($M_{Dynse}$) are trained using a combination of all three scenarios available in the PKLot: UFPR04, UFPR05, and PUCPR. The pool of classifiers consists of three models ($C_1$, $C_2$, $C_3$) trained individually on each scenario. This same protocol is applied when training the frameworks using all nine cameras from the CNR-EXT Dataset\cite{amatoEtAl2017}. In this case, the models $S$ and $M_{Dynse}$ encompass all nine scenarios, while the pool of classifiers comprises nine models ($C_1$, ..., $C_9$), each corresponding to one of the nine cameras.

The $M_{Stack}$ is trained using the given \textit{a posteriori} probabilities of each model within the pool for each training sample. For the pool comprising three PKLot models, the output will form a feature vector of 6 class probabilities for each sample, while the nine CNR-EXT models output 18 probabilities.

As detailed in Section \ref{sec:datasets}, the samples from the mentioned datasets were collected over a period of time. Therefore, to establish a fair evaluation protocol, we allocated the first 50\% of days for the training set for each scenario and reserved the remaining 50\% of days for the validation and testing set. We employed a random image drop strategy for the class with the most samples to tackle class imbalance, aligning it with the minority class in the training set. 

This approach was designed to prevent situations where a vehicle parked for an extended period or an empty parking spot with minimal changes in lighting conditions would be included in both training and test sets simultaneously, as discussed in \cite{almeidaEtAl2022}. The models were trained for 30 epochs, and the weights corresponding to the model with the best accuracy on the validation set were chosen.

Due to the impact of randomness, we conducted each experiment 10 times to avoid biased training, using different seeds to enable variations in training sets and model optimization.

The average performances from all ten measures are presented in Tables \ref{tab:acc_pklot_vs_all} and Table \ref{tab:acc_CNR_vs_all}, draw us to some conclusions. 

The frameworks trained on PKLot scenarios achieved good accuracy rates in all cross-dataset scenarios (Table \ref{tab:acc_pklot_vs_all}). Notably, a deeper representation provided by MobileNetV3 offers the best tradeoff between model complexity and accuracy for both environments. In this context, the single model approach ($S$) presents an average rate of 95.5\% when facing unseen testing environments without incurring the additional overhead caused by the pool of classifiers-based approaches. Ensemble-based strategies appear promising, mainly when working with smaller architectures. In this regard, the model $S$, by considering the Majority Vote strategy with the 3-Convolutional Layers Architecture, improved the result by 5\% when compared to the single model, reaching a result of 87\%. This is an interesting finding in situations that pose a hardware constraint.

\begin{table*}[!htbp]
\centering

\caption{Accuracies of Proposed Frameworks (\%) in Cross-Dataset Scenarios considering the PKLot Dataset for training.}
\label{tab:acc_pklot_vs_all}
\resizebox{.95\textwidth}{!}{%
\begin{tabular}{ccccccccccccc}
\hline\hline

\multirow{2}{*}{\textbf{\begin{tabular}[c]{@{}c@{}}Target Dataset / \\ Frameworks\end{tabular}}} & \multirow{2}{*}{\textbf{BarryStreet}} & \multirow{2}{*}{\textbf{NDIS}} & \multicolumn{9}{c}{\textbf{CNR-EXT}} & \multirow{2}{*}{\textbf{Average}} \\ \cline{4-12}
 &  &  & \textbf{CAM\#1} &\textbf{ CAM\#2} & \textbf{CAM\#3} & \textbf{CAM\#4} & \textbf{CAM\#5} & \textbf{CAM\#6} & \textbf{CAM\#7} & \textbf{CAM\#8} & \textbf{CAM\#9} &  \\ \hline\hline
\multicolumn{13}{c}{\textbf{3-Conv. Layers Architecture}} \\ \hline\hline
Single Model & 98.3 (0.9) & 87.1 (4.3) & 69.6 (3.1) & 74.2 (2.2) & 81.5 (1.9) & 85.8 (2.4) & 83.3 (2.2) & 85.7 (1.8) & 80.6 (2.1) & 89.2 (1.7) & 72.4 (2.8) & 82.0 (8.1) \\
Dynamic Sel & 97.7 (0.8) & 89.2 (1.6) & 73.4 (1.8) & 79.6 (2.8) & 84.6 (0.9) & 89.1 (1.3) & 86.6 (1.6) & 89.1 (1.2) & 82.4 (1.4) & 90.9 (0.7) & 80.6 (2.7) & 85.3 (6.5) \\
\textbf{Majority Vote} & {98.5 (0.7)} & {91.6 (1.3)} & {75.7 (1.7)} & {81.7 (2.9)} & {86.0 (1.3)} & {90.8 (1.4)} & {87.9 (1.7)} & {91.0 (1.2)} & {83.1 (1.6)} & {92.8 (0.8)} & {82.1 (2.6)} & \textbf{87.0 (6.2)} \\
Stacking (SVM) & 98.1 (0.8) & 87.6 (2.0) & 71.7 (2.7) & 76.7 (4.1) & 83.6 (1.5) & 88.0 (2.0) & 84.9 (2.2) & 88.1 (2.1) & 80.4 (1.8) & 90.8 (1.4) & 77.5 (3.5) & 83.9 (7.2) \\
Stacking (MLP) & 98.1 (0.8) & 87.6 (1.8) & 71.4 (2.7) & 76.4 (3.7) & 83.5 (1.5) & 87.8 (1.8) & 84.7 (2.2) & 87.9 (2.1) & 80.3 (1.8) & 90.7 (1.4) & 77.1 (3.4) & 83.7 (7.3) \\ \hline\hline
\multicolumn{13}{c}{\textbf{MobileNetV3 Architecture}} \\ \hline\hline
\textbf{Single Model} & {99.7 (0.2)} & {96.7 (0.8)} & {84.8 (1.7)} & {96.4 (0.6)} & {94.8 (0.9)} & {97.2 (0.5)} & {97.4 (0.6)} & {97.5 (0.4)} & {96.5 (0.5)} & {97.7 (0.4)} & {96.7 (0.6)} & \textbf{95.5 (4.0)} \\
Dynamic Sel & 99.1 (0.5) & 96.4 (0.7) & 83.0 (1.4) & 94.7 (1.5) & 92.8 (1.1) & 96.0 (0.5) & 96.3 (0.7) & 96.3 (0.5) & 94.8 (0.4) & 96.5 (0.5) & 95.5 (0.6) & 94.1 (4.4) \\
Majority Vote & 99.6 (0.3) & 96.8 (0.6) & 83.7 (0.9) & 96.1 (0.6) & 94.0 (0.6) & 96.7 (0.4) & 97.5 (0.3) & 97.4 (0.2) & 96.2 (0.3) & 97.4 (0.3) & 96.5 (0.4) & 95.1 (4.3) \\
Stacking (SVM) & 99.5 (0.3) & 96.4 (0.7) & 82.9 (0.9) & 95.9 (0.6) & 93.6 (0.8) & 96.7 (0.3) & 97.5 (0.3) & 97.3 (0.3) & 96.5 (0.2) & 97.5 (0.3) & 96.6 (0.4) & 95.0 (4.5) \\
Stacking (MLP) & 99.6 (0.3) & 96.6 (0.6) & 83.4 (0.9) & 96.2 (0.6) & 93.9 (0.5) & 96.8 (0.3) & 97.5 (0.3) & 97.4 (0.2) & 96.5 (0.2) & 97.6 (0.3) & 96.7 (0.4) & 95.2 (4.4) \\ \hline\hline
\multicolumn{13}{c}{\textbf{ResNet-50 Architecture}} \\ \hline\hline
\textbf{Single Model} & {98.2 (0.7)} & {94.8 (1.2)} & {82.3 (2.0)} & {95.3 (0.8)} & {93.3 (1.3)} & {96.2 (0.8)} & {97.3 (0.3)} & {97.2 (0.2)} & {96.8 (0.2)} & {96.5 (0.4)} & {96.4 (0.6)} & \textbf{94.5 (4.5)} \\
Dynamic Sel & 97.6 (0.4) & 93.8 (1.3) & 80.1 (3.2) & 93.8 (1.1) & 92.3 (1.0) & 95.9 (0.4) & 96.3 (0.5) & 96.1 (0.5) & 95.7 (0.4) & 96.2 (0.6) & 95.5 (0.6) & 93.2 (5.2) \\
Majority Vote & 98.7 (0.6) & 93.3 (0.9) & 81.0 (1.5) & 95.1 (0.9) & 92.6 (0.8) & 95.5 (0.5) & 97.0 (0.4) & 96.4 (0.4) & 96.0 (0.3) & 96.1 (0.4) & 95.9 (0.5) & 93.7 (4.9) \\
Stacking (SVM) & 98.2 (1.1) & 91.1 (1.6) & 79.4 (1.6) & 94.1 (1.1) & 91.4 (1.1) & 95.4 (0.7) & 97.0 (0.4) & 96.5 (0.5) & 95.9 (0.4) & 96.3 (0.4) & 95.6 (0.7) & 93.2 (5.3) \\
Stacking (MLP) & 98.1 (1.1) & 91.0 (1.5) & 79.3 (1.4) & 94.0 (1.0) & 91.2 (1.1) & 95.3 (0.8) & 97.0 (0.4) & 96.4 (0.5) & 95.9 (0.5) & 96.3 (0.4) & 95.6 (0.7) & 93.1 (5.3) \\ \hline\hline
\end{tabular}%
}
\end{table*}

\begin{table*}[!h]
\centering
\caption{Accuracies of Proposed Frameworks (\%) in Cross-Dataset Scenarios considering the CNR-EXT Dataset for training}
\label{tab:acc_CNR_vs_all}
\resizebox{.60\textwidth}{!}{%
\begin{tabular}{ccccccc}
\hline\hline

\multirow{2}{*}{\textbf{\begin{tabular}[c]{@{}c@{}}Target Dataset / \\ Frameworks\end{tabular}}} & \multirow{2}{*}{\textbf{BarryStreet}} & \multirow{2}{*}{\textbf{NDIS}} & \multicolumn{3}{c}{\textbf{PKLot}} & \multirow{2}{*}{\textbf{Average}} \\ \cline{4-6}
 &  &  & \textbf{UFPR04} & \textbf{UFPR05} & \textbf{PUCPR} &  \\ \hline\hline
\multicolumn{7}{c}{\textbf{3-Conv. Layers Architecture}} \\ \hline\hline
Single Model & 98.3 (0.6) & 98.3 (0.5) & 59.8 (3.7) & 67.4 (2.2) & 79.7 (2.9) & 80.7 (17.5) \\
Dynamic Sel & 94.8 (2.0) & 97.5 (1.0) & 64.8 (5.5) & 66.5 (1.7) & 81.8 (2.6) & 81.1 (15.3) \\
\textbf{Majority Vote} & 98.1 (0.9) & 99.2 (0.4) & 64.6 (5.4) & 64.8 (0.8) & 83.4 (2.8) & \textbf{82.0 (17.0)} \\
Stacking (SVM) & 98.2 (0.5) & 99.3 (0.2) & 61.4 (3.5) & 64.4 (0.4) & 82.2 (2.2) & 81.1 (18.0) \\
Stacking (MLP) & 97.4 (2.3) & 99.3 (0.3) & 59.9 (3.4) & 64.5 (0.9) & 81.3 (2.5) & 80.5 (18.2) \\
\hline\hline
\multicolumn{7}{c}{\textbf{MobileNetV3 Architecture}} \\ \hline\hline
Single Model & 99.2 (0.8) & 98.4 (0.5) & 73.4 (6.1) & 81.5 (4.3) & 91.1 (3.5) & 88.7 (11.1) \\
Dynamic Sel & 99.0 (0.3) & 97.0 (1.1) & 74.9 (6.2) & 82.2 (2.8) & 93.4 (2.4) & 89.3 (10.3) \\
\textbf{Majority Vote} & 99.3 (0.2) & 98.5 (0.4) & 78.0 (4.5) & 82.4 (3.2) & 92.3 (1.2) & \textbf{90.1 (9.6)} \\
Stacking (SVM) & 99.2 (0.3) & 99.0 (0.2) & 73.2 (3.3) & 81.1 (2.8) & 90.7 (1.7) & 88.6 (11.4) \\
Stacking (MLP) & 99.1 (0.2) & 99.1 (0.3) & 72.2 (4.1) & 80.8 (3.0) & 91.2 (1.7) & 88.5 (11.8) \\
\hline\hline
\multicolumn{7}{c}{\textbf{ResNet-50 Architecture}} \\ \hline\hline
Single Model & 97.6 (1.4) & 97.5 (1.1) & 67.7 (5.6) & 80.0 (3.3) & 95.2 (2.0) & 87.6 (13.3) \\
Dynamic Sel & 97.2 (1.5) & 96.0 (1.7) & 64.2 (6.6) & 78.0 (5.1) & 94.9 (1.2) & 86.1 (14.5) \\
\textbf{Majority Vote} & 98.9 (0.2) & 98.4 (0.4) & 73.1 (4.1) & 77.9 (2.5) & 95.4 (0.6) & \textbf{88.7 (12.3)} \\
Stacking (SVM) & 97.9 (0.7) & 99.3 (0.2) & 54.1 (4.0) & 67.4 (2.4) & 85.9 (4.5) & 80.9 (19.7) \\
Stacking (MLP) & 98.2 (0.5) & 99.3 (0.2) & 54.2 (3.2) & 67.7 (2.0) & 88.1 (1.8) & 81.5 (19.8) \\
\hline\hline

\end{tabular}%
}
\vspace{-2mm}
\end{table*}

When considering the results in Table \ref{tab:acc_CNR_vs_all}, we can see a sharp drop in the best average accuracy achieved, with the MobileNetV3 architecture reaching a result of 90.1\% when considering the Majority Voting Strategy. We hypothesize that even though the CNRPark-EXT dataset has nine cameras, they all capture the same scenario simultaneously, leaving unchanged environmental issues such as lighting, weather, and background textures for all cameras. This lack of diversity deteriorates the model generalization. The PKLot Dataset, in contrast, offers greater variety with two distinct scenarios (UFPR and PUC). Although PKLot has two cameras for the same scenario (UFPR04 and UFPR05), the images were taken on different days.

Notice that by using a test protocol that considers four distinct datasets and using a single dataset for training in a cross-dataset scenario, we created a more realistic scenario when compared with most state-of-the-art works (e.g., in \cite{HochuliEtAl2022-Annotation} the authors considered a cross-parking lot scenario, nevertheless a single dataset was used). Nonetheless, our results align with the state-of-the-art, and we can conclude that:

\textbf{RQ1: How accurate are existing deep learning models when applied in a cross-dataset scenario?} We achieved averaged results of 95\% when considering a deep learning model, although the training must consider diverse scenarios to make it possible for the model to generalize its representations.

\textbf{RQ2: Regarding different architectures and ensemble strategies, which framework is the most suited for cross-dataset scenarios?}  Towards a global framework, a Single Model trained with the MobileNetv3 showed the best results and tradeoff between the number of parameters and achieved results. The Majority Voting strategy may slightly improve the results when the training data is not diverse enough or the classifier has a weak generalization.

\section{Conclusion}\label{sec:conclusion}
   
The critical discussion presented in Section \ref{sec:experiments} has given us valuable insights to answer our research questions correctly. First, we demonstrate that the MobileNetV3 architecture is well-suited for parking spot classification across scenarios, simulating real-world deployments where no fine-tuning is a constraint (RQ1). In this context, a single global model approach, trained over diverse scenarios, can better generalize the task and outperform any of the proposed ensemble frameworks, achieving an average rate of 95\%. However, ensembles could still yield some benefits in situations requiring smaller architectures, such as when hardware is a constraint (RQ2). Finally, the PKLoT dataset has shown better data diversity to train a generalist framework.    

In future work, other ensemble methods (i.e., heterogeneous pools) and the performance drop in the CNR-EXT dataset are a matter of deep investigation.

\section*{Acknowledgment}
This work has been supported by the Brazilian National Council for Scientific and Technological Development (CNPq) – Grant 405511/2022-1

\bibliographystyle{IEEEtran}
\bibliography{paper}

\end{document}